\documentclass{article}

\usepackage{PRIMEarxiv}

\usepackage[utf8]{inputenc} 
\usepackage[T1]{fontenc}    
\usepackage{hyperref}       
\usepackage{url}            
\usepackage{booktabs}       
\usepackage{amsfonts}       
\usepackage{nicefrac}       
\usepackage{microtype}      
\usepackage{lipsum}
\usepackage{fancyhdr}       
\usepackage{graphicx}       
\graphicspath{{media/}}     
\usepackage{orcidlink}
\usepackage{tikz}
\usepackage{pgfplots}
\usepackage{subcaption}
\usepackage{float}
\usepackage{kotex}

\title{Efficacy of Utilizing Large Language Models to Detect Public Threat Posted Online
}

\author{
  \orcidlink{0009-0004-5775-2550} Taeksoo Kwon \\
  Algorix Convergence Research Office \\
  New York, New York \\
  \texttt{henryk@research.algorix.io} \\
   \And
  \orcidlink{0009-0005-7375-5492} Connor Kim \\
  Centennial High School \\
  Corona, California \\
  \texttt{connorkusa@gmail.com} \\
}

\begin{document}
\maketitle

\begin{abstract}
This paper examines the efficacy of utilizing large language models (LLMs) to detect public threats posted online. Amid rising concerns over the spread of threatening rhetoric and advance notices of violence, automated content analysis techniques may aid in early identification and moderation. Custom data collection tools were developed to amass post titles from a popular Korean online community, comprising 500 non-threat examples and 20 threats. Various LLMs (GPT-3.5, GPT-4, PaLM) were prompted to classify individual posts as either “threat” or “safe.” Statistical analysis found all models demonstrated strong accuracy, passing chi-square goodness of fit tests for both threat and non-threat identification. GPT-4 performed best overall with 97.9\% non-threat and 100\% threat accuracy. Affordability analysis also showed PaLM API pricing as highly cost-efficient. The findings indicate LLMs can effectively augment human content moderation at scale to help mitigate emerging online risks. However, biases, transparency, and ethical oversight remain vital considerations before real-world implementation.
\end{abstract}

\keywords{Large Language Model \and Content Moderation \and Public Threat}

\section{Introduction}
The evolution of online communities has propelled the amalgamation of diverse perspectives and ideologies, fostering an environment where information flows unimpeded across geographical boundaries \cite{know_col}. However, the cloak of anonymity veils the identities of individuals, allowing for the unrestrained expression of thoughts and actions. This anonymity, while liberating in some aspects, has unfurled a tapestry of challenges, manifesting prominently in the form of group polarization \cite{internet_inter}. The phenomenon of group polarization, accentuated in the digital realm, intensifies ideological extremities within communities, creating echo chambers that reinforce existing beliefs and amplify radical perspectives. Moreover, the ease of disseminating harmful content in these spaces, fueled by the shield of anonymity, presents a critical societal challenge \cite{internet_anom}.

Incidents, like the distressing sequence of copycat crimes following the Sillim subway station tragedy, underscore the alarming consequences of unchecked propagation of harmful ideas and actions within online platforms \cite{copycat}. The initial act of violence echoed into a haunting symphony of fear as a slew of imitative offenses gripped the nation, reverberating not only in physical spaces but also pervading the digital realm \cite{police_releases_sillim}. Due to its widespread popularity, online platforms like DC Inside transformed into channels for unfiltered hate speech and threats \cite{kor_ranking_online}. This surge intensified the difficulty of content moderation, amplifying the constraints posed by human limitations as outlined in prevailing online content governance models. Concurrently, it escalated societal tensions, manifesting through a barrage of virtual and physical threats directed at random individuals \cite{fears_grow_copycat}. Traditional content moderation methods, predominantly reliant on human intervention, grapple with the daunting task of sifting through voluminous user-generated content. Platforms, bound by community guidelines, strive to uphold standards by curtailing the dissemination of objectionable material. However, the sheer scale and diversity of content pose significant hurdles, stretching the capabilities of human moderators to their limits \cite{human_cost_online}.

In light of these challenges, the exploration of artificial intelligence, particularly large language models (LLM), emerges as a beacon of hope. These AI-driven mechanisms possess the potential to augment existing content moderation practices by leveraging sophisticated algorithms capable of swiftly analyzing, categorizing, and flagging problematic content. The ability of AI models to process vast amounts of data at unparalleled speeds offers a glimmer of optimism in addressing the inadequacies of human-centric moderation strategies.

\section{Literature Review}
This literature review delves into the evolving landscape of LLM applications within Natural Language Processing (NLP). Recent research emphasizes the efficacy of autoregressive models, notably GPT, LLaMa, and Alpaca. The advent of GPT precipitated a surge in interest surrounding LLMs in generative AI, leading to varied developments in the field. Notably, the open-sourcing of LLaMa facilitated the customization and refinement of LLMs by developers \cite{llama_opensource}.

Integrating LLMs into real-world applications necessitates consideration of diverse components, such as plugins, frameworks, and agents. These elements enable the incorporation of external data and functionalities, language assistance, and seamless integration interfaces, culminating in comprehensive implementations. Applications in sectors like healthcare and science, such as DNA-oriented LLMs, exemplify the diverse applications of LLMs \cite{llm_integrations}.

\subsection{Recent Trends in Online Threats in America}
The Pew Research Center conducted research in 2021, revealing that 41 percent of Americans have personally experienced some form of online harassment, ranging from common to severe comments \cite{online_harassment}. While the trend of online threats stayed consistent since 2017, it is quite evident that it has intensified over the years. These threats can be identified from least to greatest in terms of offense, offensive name-calling, purposeful embarrassment, stalking, sustained harassment, sexual harassment, and physical threats.

\subsection{Trends in South Korea: Tragedy at Sillim Station}
Regarding the research, South Korea has experienced an abnormal growth in online threats, some even correlating to real life. The Sillim Station stabbings played a major role in the expansion of threats in Korea, contributing approximately 430 murder posts in that month alone after the incident \cite{sillim_threats}. Seohyeon Station, Gwanaksan Park, Hapjeong Station, Gwangmyeong Station, and other places, Sillim being the most notable, were all locations that contained certain threats according to the posts in Korea by their communities. Because of the escalation of the situation, Korea reinforced its security by enforcing police to track the people who have posted threats and take information from citizens about possible crimes that will be committed \cite{namu_silim_incidents}.

Overall, the impersonation trend enabled the diffusion and normalization of more threatening rhetoric relating to the Sillim Station case through covert means online. This highlighted the rise in popularity of the emergence of security in online communities and helped provide alternative solutions such as Terrorless which was created to map out locations where possible crimes may be committed through user reports and online posts \cite{jung_2023}.

\subsection{The Rise of Large Language Models: What are They?}
LLM (Large Language Model), is an advanced deep learning model for natural language processing that is trained on vast amounts of text data. Some notable LLMs include BERT, GPT-3, GPT-4, and XLNet developed by teams at Google, OpenAI, Anthropic, and others. The goal of an LLM is to understand and generate human language at a high level through massive computing power and datasets. 

These language models are now being utilized across many industries and fields to augment human capabilities. For example, GitHub has created their own LLM, CoPilot X to help assist and support developers in programming and fixing errors \cite{copilot_X}. In customer support, chatbots powered by LLMs can answer common queries automatically around the clock \cite{chatbots}. LLMs are also used for content creation, science, and research by analyzing papers and data. Furthermore, they assist professionals in legal, medical, and other domains by reviewing documents, conducting research, and answering questions to accelerate their work.

\subsection{Popularity and Affordability of Large Language Models}
OpenAI's gpt-3.5-turbo-1106 and gpt-4, alongside PaLM API for Google Bard, stand as prominent publicly available LLMs \cite{popular_large_llm}. When considering the cost aspect, gpt-3.5-turbo-1106 is priced at \$0.001 per 1K tokens for input and \$0.002 per 1K tokens for output, while gpt-4 costs \$0.03 per 1K tokens for input and \$0.06 per 1K tokens for output \cite{gpt_pricing}. On the other hand, PaLM 2 for Chat (chat-bison) operates at \$0.00025 per 1K characters for input and \$0.0005 per 1K characters for output \cite{palm_pricing}.

These varying pricing structures make PaLM 2 for Chat notably more affordable in comparison to both gpt-3.5-turbo-1106 and gpt-4, providing an attractive option for certain applications due to its lower cost per token/character. Despite this, the popularity of these models is influenced by factors beyond just affordability, including performance, capabilities, and accessibility, contributing to their diverse usage across different contexts within the LLM landscape \cite{llms_stats}.

\subsection{Cases of Using LLM to Moderate Content}
OpenAI researchers have begun applying generative models to the application of online content moderation. As described in a blog post, their method utilizes GPT-4 to assist in developing and continually refining platform-specific content policies on issues like hate speech, abuse, and threats in a highly automated and scalable manner \cite{weng_goel_vallone}. Through an iterative process of policy drafting, example curation, and model feedback, they aim to speed up the traditionally lengthy process of policy evolution from months to just hours.

A similar case was led by a team of researchers who investigated LLMs and Content Moderation and found that LLMs can be effective in rule-based content moderation and toxic detection \cite{kumar_abuhashem_durumeric_2023}. The researchers tested LLMs on rule-based community moderation and toxic content detection and found that LLMs can be effective for rule-based moderation and outperform existing toxicity classifiers. However, they also found that the increase in model size only provides a marginal benefit for toxicity detection. The researchers acknowledge that their results may not extend to other types of moderation and that the cost of LLMs is currently high. Resulting that while LLMs show promise, more research is needed to balance performance with cost. The research also includes a case study on the subreddit r/worldnews, highlighting the errors made by the LLM in moderation decisions. Overall, their research provides a tempered but optimistic view of using LLMs in content moderation and suggests avenues for future research.

With continued experimentation integrating techniques like distillation and active learning, LLMs may help alleviate some of the mental burden on human moderators while also enabling faster responses to emerging online risks. However, limitations around unwanted biases potentially introduced during pretraining also underscore the need for careful oversight and model validation as these systems grow in real-world impact.

\subsection{Limitations of Human Content Moderation}
While this study focuses on evaluating the technical capabilities of large language models for content moderation, it's worth noting some of the inherent limitations of relying solely on human moderators. At massive scales, human content moderation can be an inefficient, costly, and imperfect process \cite{ai_human_moderation}.

Human content moderation at large scales faces significant challenges that make it an inefficient and imperfect solution on its own. Moderators are susceptible to factors like fatigue, stress, and burnout from constant exposure to harmful or disturbing content \cite{content_traum}. This can negatively impact judgment and consistency over time. Individual biases, cultural differences, and personal interpretations of guidelines also introduce elements of unpredictability and subjectivity. As online communities continue growing exponentially, it becomes impossible for human reviewers alone to keep up with demanded capacities \cite{human_mod}.

A research conducted by Petter Törnberg explains the accuracy and reliability of ChatGPT, an LLM, compared to human classifiers. The research compares the performance of ChatGPT with crowd workers on MTurk and expert classifiers. It is found that ChatGPT outperforms individual human classifiers \cite{törnberg_4AD}

Automated techniques using AI models have the potential to supplement human moderation by helping address some of these limitations at scale. Overall, the limitations indicate a need for supplemental automated techniques.

\subsection{Ethical Considerations}
Employing AI models for content analysis raises important ethical questions around privacy, accountability, and unintended consequences. As these systems continue expanding in impact and autonomy, key considerations include transparency around how decisions are made, access to appeal channels, and validation of model fairness \cite{ai_concerns}.
    
By training AI systems on data generated by humans, there is a potential risk of replicating and even amplifying the possible bias problems. Different groups could potentially be treated unfairly or censored at disproportionate rates based on characteristics like gender, race, political views, etc \cite{ethical_ai}. Content moderation systems must be developed and evaluated using intersectional approaches to proactively identify and mitigate against such biases. Without adequate measures, biases in automated censorship could end up promulgating wider societal inequalities and exclusion online \cite{ai_pros_cons}.

Relying too heavily on AI systems for content moderation could potentially lead to many human moderators losing their jobs. This could have negative economic and social impacts if the transition is not thoughtfully managed. The role of humans also helps ensure appropriate oversight and governance over these important decision processes. By delegating broad moderation powers to automated systems, it effectively extends its authority over information flows and discussions online. There are risks of over-zealous or biased censorship if models are not carefully designed, overseen, and calibrated. It's also difficult to understand and challenge specific AI decisions, undermining principles of transparency. Overall governance structures would need careful consideration to avoid such unintended consequences \cite{ai_concerns}.

\subsection{Future Directions} 
The exploration of artificial intelligence, particularly large language models (LLMs), for online content moderation presents a promising avenue for the future. As technology continues to advance, several key directions emerge for the integration and enhancement of LLMs in the realm of content moderation.

\section{Methodology}
This study examines the effectiveness of large language models in detecting public threats within an online community. To curate a diverse dataset inclusive of genuine threat posts, strict adherence to the community's scraping policies, as outlined in robots.txt, was prioritized. A custom HTTP tool was crafted to systematically gather and analyze data via OpenAI's official GPT models accessed through their public API. Additionally, the same HTTP tool was utilized to send requests to the PaLM API, which operates Google Bard, enhancing the breadth of model analysis \cite{Kwon_Threat_Post_Detection}. This method ensures ethical data collection while exploring the capacity of AI models to identify threats within varied online forums.

\subsection{Research Design}
The research design in this study is centered around experimentally exploring the feasibility of employing large language models (LLMs) for content moderation within online communities. The primary objective is to assess the capacity of these models, including GPT variants and others, to identify potentially threatening content among a diverse range of posts. However, it's important to clarify that this investigation doesn't specifically delve into the affordability or cost-effectiveness associated with using LLMs for content moderation purposes. The focus remains on evaluating the effectiveness of these models and their potential utility in mitigating threats, rather than on economic considerations related to their implementation.

\subsection{Data Collection and Analysis}
The data collection phase involved extracting 500 post titles from the renowned online platform "DC Inside"\footnote{https://www.dcinside.com/}, specifically from the "실시간 베스트 갤러리" (Real-time Best Gallery). A specialized scraping tool was meticulously used to exclude any posts containing public threat content from this dataset \cite{Kwon_DC_Inside_Website}. Additionally, an auxiliary dataset of 20 post titles sourced from verified threat post content was compiled. To simulate real-world scenarios, 40 non-threatening post titles were randomly selected for each attempt alongside 10 threatening titles, and their sequence was deliberately shuffled. This composite dataset underwent analysis using various large language models, including OpenAI's gpt-3.5-turbo-1106 and gpt-4, as well as PaLM API's chat-bison.

\begin{figure}[H]
\begin{itemize}
    \item 10분뒤 천호동 디스코팡팡에서 폭탄테러,ㅋㅂㄹ 한다
    \item 여의도에서 칼부림한다
    \item 수요일날 드론으로 신촌역인근 공중서 펜타닐살포예정
    \item 다음주 화요일 이수역에서 10명 칼로 찔러 죽일거다.
    \item 오늘자 약간 맛이간 삼수생 칼부림 예고...
    \item 인천공항 폭탄테러 8시 예정
    \item 오늘 오후 7시에 강남역에서 100명 죽일 예정
    \item 오늘 밤 11시에 이대역에서 칼부림 예정입니다
    \item 내가 구라안치고 일요일날 칼들고 개좆슼갈들 전부 찔러 죽일거임.
    \item 다음주 수요일 5시에 강남역에서 경찰관10명 칼로 쳐죽일거다
    \item \dots
\end{itemize}
\caption{Example Public Threat Post Titles Collected \cite{namu_silim_incidents}}
\label{fig:ex-threats}
\end{figure}

Subsequent statistical analysis, specifically the chi-square goodness of fit test at a general significance level of 0.05, was conducted to determine the suitability of employing these LLMs, considering their performance in content moderation \cite{franke_ho_christie_2011}. This comprehensive evaluation aimed to assess the models' proficiency in distinguishing between threatening and non-threatening content within the dynamic landscape of an online community, taking into account the nuanced capabilities of various LLMs.

\subsection{Prompt Engineering}
The experimental design encompassed the repetition of this meticulous process for enhanced reliability and robustness \cite{prompt_eng}. Iterating this procedure 25 times reinforced the rigor of the assessment. Through each iteration, the 40 safe and 10 threatening post titles were individually funneled through the designated APIs. This iterative approach not only amplified the sample size but also provided a comprehensive overview of the models' consistency and reliability in accurately classifying diverse content. Consequently, this methodological strategy bolstered the statistical significance of the findings, affording a more nuanced understanding of the models' performance and their ability to effectively discern threat-related content within an online community setting. The base prompt used for this purpose is as follows:

\begin{quote}
    Beda is a content moderator working for a Korean online community website. Beda is given a post title in Korean to review. Beda judges the post as either safe or unsafe based on whether it contains content posing a public threat. Beda defines a public threat as advance notices that perpetrators upload before committing terror in public places. Beda must be considerate to keep the public safe. Beda can only reply in either 'threat' or 'safe', based on whether the text is a public threat or not. Beda does not translate the text into other languages. Beda should only reply in one word. Beda does not add any description about the message. You are Beda.
\end{quote}

\subsection{Evaluation Criteria}
The assessment criteria for content moderation employing large language models primarily revolves around accuracy. The evaluation methodology relies on the chi-square test, utilizing the resulting test statistic to establish decision criteria \cite{franke_ho_christie_2011}. Should the computed statistic exceed the critical value, predetermined by a significance level of 0.05, rejecting the null hypothesis indicates a significant deviation of the data from the anticipated distribution. This deviation suggests a lack of significant accuracy in the model's ability to identify threats within the content being evaluated, supporting the hypothesis that the model might not possess substantial capabilities for threat detection.

Conversely, the inability to reject the null hypothesis indicates that the data conforms to the specified distribution. This conformity signifies that the model demonstrates notable accuracy in detecting threats within the dataset. This statistical framework provides a decisive metric for assessing the model's performance in discerning threatening content within the curated dataset.

\subsection{Ethical Considerations}
The methodology adhered meticulously to ethical considerations throughout the data collection process. Compliance with ethical guidelines was paramount, adhering strictly to the directives outlined in the robots.txt file and adhering to DC Inside's Terms of Service (TOS) \cite{zachary_mark_2018}\cite{dcinside_tos}. These measures ensured respect for the online community's guidelines and policies, preserving the integrity of data acquisition while safeguarding the platform's regulations. Furthermore, an anonymous approach was adopted during post-collection to uphold user privacy and confidentiality. This anonymity shielded individual identities and protected user information, underscoring a commitment to ethical data-handling practices throughout the research process.

\section{Results}
The experimentation was structured around a series of 30 distinct trials aimed at rigorously assessing the performance and consistency of the developed methodology. Each trial offered unique insights into the efficacy of the approach in detecting threats within online post titles. Notably, all code utilized throughout the experimentation process has been meticulously documented and made available on GitHub \cite{Kwon_DC_Inside_Website}\cite{Kwon_Threat_Post_Detection}. This comprehensive repository houses the tools, scripts, and methodologies employed, ensuring transparency, reproducibility, and accessibility for further scrutiny and replication of the study's procedures.

\subsection{Data Collected}

\begin{table}[H]
\centering
\begin{tabular}{|c|c|c|c|c|c|}
\hline
\textbf{Model Name} & \textbf{Non-threats Judged Correctly} & \textbf{Threats Judged Correctly} \\
\hline
gpt-3.5-turbo-1106 & 32.6 & 9.92 \\
gpt-4 & 39.16 & 10 \\
chat-bison (PaLM2) & 38.6 & 10 \\
\hline
\end{tabular}
\caption{Raw Data Collected from Each Model}
\label{tab:data-collection}
\end{table}

\subsection{Performance Analysis}

\begin{figure}[H]
\centering
\begin{tikzpicture}
\begin{axis}[
    ybar,
    xlabel={Model Name},
    ylabel={Non-threats Judged Correctly (\%)},
    ymin=0,
    ymax=100,
    width=12cm,
    height=8cm,
    symbolic x coords={gpt-3.5-turbo-1106, gpt-4, chat-bison},
    xtick=data,
    nodes near coords,
    nodes near coords align={vertical},
    bar width=12pt,
    every node near coord/.append style={yshift=5pt},
]
\addplot+[error bars/.cd, y dir=both, y explicit]
    coordinates {
        (gpt-3.5-turbo-1106, 81.5) +- (0, 3.18)
        (gpt-4, 97.9) +- (0, 0.28)
        (chat-bison, 96.5) +- (0, 0.99)
    };
\end{axis}
\end{tikzpicture}
\caption{Non-threat Accuracy Comparison}
\label{fig:non-threat-graph}
\end{figure}

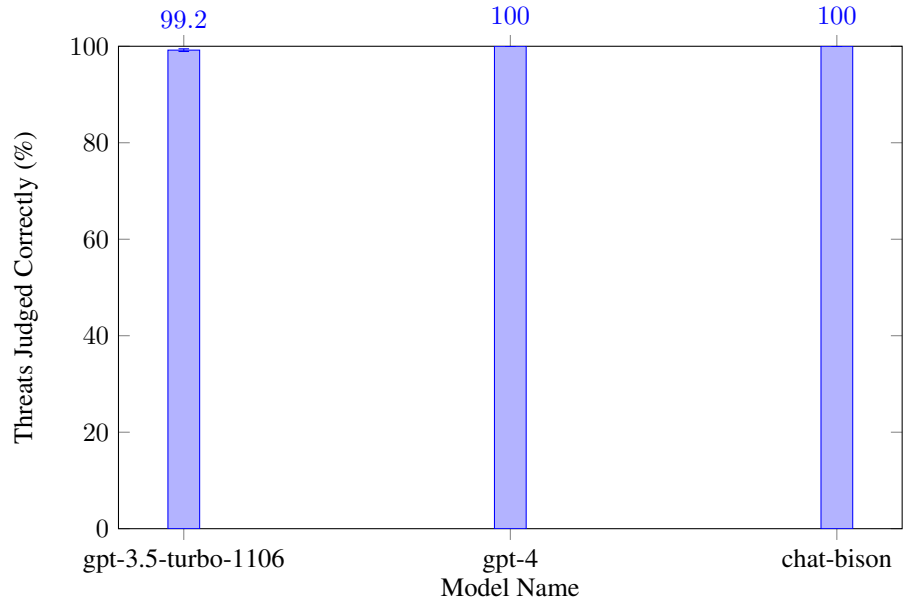
\begin{figure}[H]
\centering
\begin{tikzpicture}
\begin{axis}[
    ybar,
    xlabel={Model Name},
    ylabel={Threats Judged Correctly (\%)},
    ymin=0,
    ymax=100,
    width=12cm,
    height=8cm,
    symbolic x coords={gpt-3.5-turbo-1106, gpt-4, chat-bison},
    xtick=data,
    nodes near coords,
    nodes near coords align={vertical},
    bar width=12pt,
    every node near coord/.append style={yshift=5pt},
]
\addplot+[error bars/.cd, y dir=both, y explicit]
    coordinates {
        (gpt-3.5-turbo-1106, 99.2) +- (0, 0.28)
        (gpt-4, 100) +- (0, 0)
        (chat-bison, 100) +- (0, 0)
    };
\end{axis}
\end{tikzpicture}
\caption{Threat Accuracy Comparison}
\label{fig:threat-graph}
\end{figure}

\subsection{Chi-Square Tests}
Chi-square goodness of fit tests were conducted to evaluate the performance of each model in detecting public threats \cite{franke_ho_christie_2011}. The null hypothesis ($H_0$) assumed that there would be no difference between the observed and expected values, indicating that the model is capable of accurately detecting public threats. On the other hand, the alternative hypothesis ($H_a$) posited that a significant difference would exist between the observed and expected values, suggesting that the model is incapable of effectively detecting public threats. The chi-square tests were employed to assess whether the observed results deviated significantly from the expected results, providing insights into the model's performance in identifying potential public threats.

\begin{equation}
    \chi^2 = \sum \frac{{(O_i - E_i)^2}}{{E_i}}
\end{equation}

\subsubsection{gpt-3.5-turbo-1106}
The calculated chi-square value ($\chi^2$ = 1.370) with 1 degree of freedom did not exceed the critical value of the chi-square distribution at a significance level of 0.05 ($\chi^2$ distribution = 3.84). Therefore, the test does not provide sufficient evidence to reject the null hypothesis of a good fit between the observed and expected values.

\begin{equation}
    \chi^2 = \frac{{(32.6 - 40)^2}}{{40}} + \frac{{(9.92 - 10)^2}}{{10}} \approx 1.370
\end{equation}

\begin{table}[H]
\centering
\begin{tabular}{|c|c|c|}
\hline
& \textbf{Non-threat Correct} & \textbf{Threat Correct} \\
\hline
Observed & 32.6 & 9.92 \\
\hline
Expected & 40 & 10 \\
\hline
\end{tabular}
\caption{Chi-Square Test for gpt-3.5-turbo-1106}
\label{tab:data-collection}
\end{table}

\subsubsection{gpt-4}
The calculated chi-square value ($\chi^2$ = 0.018) with 1 degree of freedom did not exceed the critical value of the chi-square distribution at a significance level of 0.05 ($\chi^2$ distribution = 3.84). Consequently, the test does not provide enough evidence to reject the null hypothesis of a good fit between the observed and expected values.

\begin{equation}
    \chi^2 = \frac{{(39.16 - 40)^2}}{{40}} + \frac{{(10 - 10)^2}}{{10}} \approx 0.018
\end{equation}

\begin{table}[H]
\centering
\begin{tabular}{|c|c|c|}
\hline
& \textbf{Non-threat Correct} & \textbf{Threat Correct} \\
\hline
Observed & 39.16 & 10 \\
\hline
Expected & 40 & 10 \\
\hline
\end{tabular}
\caption{Chi-Square Test for gpt-4}
\label{tab:data-collection}
\end{table}

\subsubsection{chat-bison}
The calculated chi-square value ($\chi^2$ = 0.049) with 1 degree of freedom did not exceed the critical value of the chi-square distribution at a significance level of 0.05 ($\chi^2$ distribution = 3.84). Hence, the evidence from the test is insufficient to reject the null hypothesis of a good fit between the observed and expected values.

\begin{equation}
    \chi^2 = \frac{{(38.6 - 40)^2}}{{40}} + \frac{{(10 - 10)^2}}{{10}} \approx 0.049
\end{equation}

\begin{table}[H]
\centering
\begin{tabular}{|c|c|c|}
\hline
& \textbf{Non-threat Correct} & \textbf{Threat Correct} \\
\hline
Observed & 38.6 & 10 \\
\hline
Expected & 40 & 10 \\
\hline
\end{tabular}
\caption{Chi-Square Test for chat-bison}
\label{tab:data-collection}
\end{table}

\subsection{Discussion}
Across the tested models, a consistent trend emerged regarding their performance in threat detection. All exhibited commendable accuracy levels, particularly in identifying potential threats within content. Notably, GPT-4 demonstrated the highest accuracy in both non-threat (97.9\%) and threat (96.5\%) identification, showcasing its proficiency in this domain.

Conversely, GPT-3.5 displayed comparatively lower accuracy scores for both non-threat (81.5\%) and threat (99.2\%) identification tasks. This observation highlights certain limitations within the GPT-3.5 model concerning its effectiveness in content moderation, signifying areas for potential improvement.

\begin{figure}[H]
    \centering
    \includegraphics[width=0.5\linewidth]{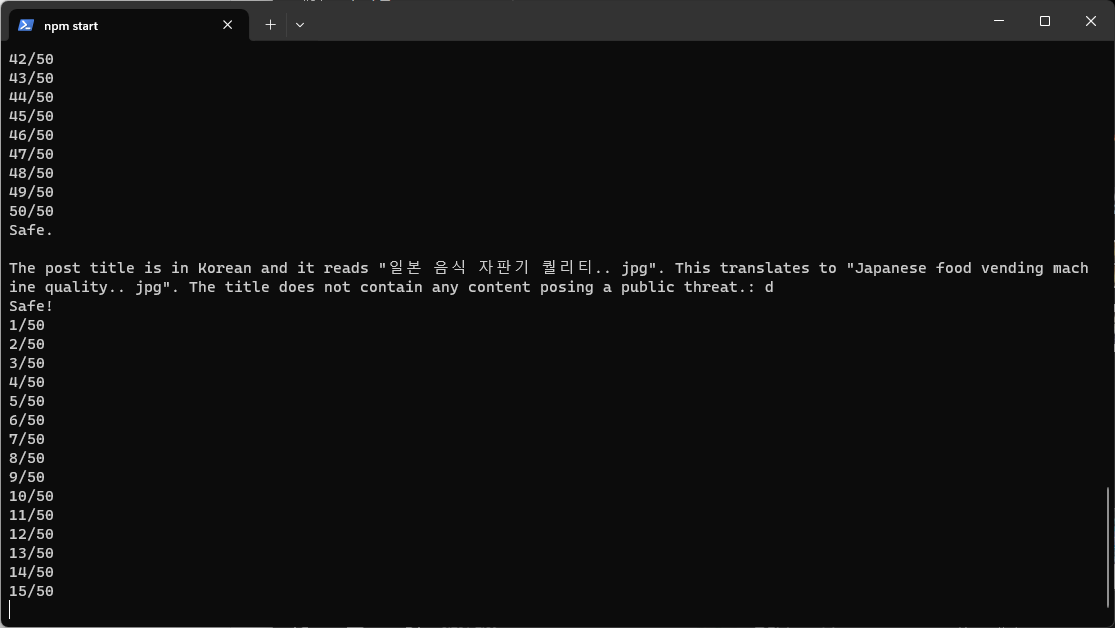}
    \caption{Screenshot of Manually Interpreting PaLM API's Result}
    \label{fig:palm-manual}
\end{figure}

Among the factors influencing model selection, affordability emerged as a crucial consideration. The cost efficiency analysis indicated that the PaLM API's Chat-Bison presented an economically favorable option \cite{palm_pricing}. However, while it showcased promise, challenges arose during the retrieval of responses in a specific format from Chat-Bison. Its tendency to occasionally provide nuanced responses, rather than a clear 'threat' or 'safe' classification as per the prompt, necessitated manual judgment for threat assessment based on the content of the response \ref{fig:palm-manual}. Enhancements in this aspect could be achieved by tailoring prompts specifically for each model, taking into account their distinct design characteristics \cite{llm_oph}.

It's imperative to note that the dataset used in this research was collected in Korean, emphasizing the potential for different outcomes when employing these models with English content \ref{fig:ex-threats}. This distinction highlights the importance of language and context in training and utilizing AI models for threat detection, suggesting that results might vary significantly when applied to distinct linguistic and cultural environments \cite{rust_pfeiffer_vuli_ruder_gurevych_2021}.

\section{Conclusions}
The comprehensive evaluation of various models for content moderation has provided invaluable insights into their effectiveness and applicability. Across the board, all models successfully passed the rigorous chi-square test at a significance level of 0.05, affirming their statistical viability for real-world content moderation applications. This outcome underscores their potential practical utility.

Ultimately, this evaluation illuminates the strengths and considerations surrounding various models for content moderation. It underscores the significance of a balanced assessment encompassing accuracy, affordability, and interpretability when selecting models for real-world content moderation scenarios. Moreover, this research serves as a foundation, illustrating the potential pathways for leveraging Large Language Models in future content moderation strategies. As advancements continue in model development and implementation strategies, these findings pave the way for more efficient and adaptable content moderation solutions, offering insights into harnessing the capabilities of LLMs for effective moderation in diverse online environments.

\section{Acknowledgements}
We acknowledge the indispensable contributions of the online community, DC Inside, for their assistance in collecting safe post titles data, and the editors of Namuwiki for their efforts in gathering threat posts, ultimately benefiting our study and the wider community.

Furthermore, special thanks to Dr. Hojin Moon, a luminary in the field of Statistics and a revered professor at California State University Long Beach for his invaluable support and meticulous review of our statistical analysis method.

\bibliographystyle{unsrt}  
\bibliography{references}

\end{document}